\documentclass{article}
\usepackage{graphicx}
\usepackage{amsmath}
\usepackage{amssymb}
\usepackage{booktabs}
\usepackage{caption}
\usepackage{wrapfig}
\usepackage{multirow}
\usepackage{makecell}
\usepackage{array}
\usepackage{placeins}
\usepackage[preprint]{corl_2026}

\title{Acting While Understanding: Asynchronous Semantic-Action Decoupling for Real-Time Vision-Language-Action Models}

\author{
  \normalfont
  Shenhao Yan\textsuperscript{1,2,*} \quad
  Ge Wang\textsuperscript{2,3,*} \quad
  Qi Liu\textsuperscript{2} \quad
  Weilin Meng\textsuperscript{2} \quad
  Jiahao Yang\textsuperscript{2} \\
  Chengsi Yao\textsuperscript{2} \quad
  Fan Feng\textsuperscript{2} \quad
  Xiaoguang Ma\textsuperscript{1,\textdagger} \quad
  Yiming Zhao\textsuperscript{2,\textdagger} \quad
  Yatong Han\textsuperscript{2,\textdagger} \\
  \small
  \textsuperscript{1} Northeastern University \\
  \textsuperscript{2} Ising AI \\
  \textsuperscript{3} CUHK-Shenzhen
}

\begin{document}

\maketitle
\begingroup
\renewcommand{\thefootnote}{\fnsymbol{footnote}}
\footnotetext[1]{Equal contribution.}
\footnotetext[2]{Corresponding authors. Contact: \texttt{rstanten@alumni.stanford.edu}.}
\endgroup

\begin{abstract}
Vision-Language-Action models (VLAs) have demonstrated strong task understanding and generalization in robotic manipulation, yet the high computational cost of full-model inference limits their deployment in low-latency, high-frequency closed-loop control. We propose an asynchronous semantic-action decoupling framework that separates semantic understanding from action generation along the internal semantic-action interface of existing VLAs, without redesigning the vision-language backbone or introducing an external planner. A low-frequency understanding module asynchronously updates reusable semantic conditions, while a high-frequency action module continuously outputs control actions without repeatedly invoking the full model. To mitigate the temporal mismatch between stale semantics and the current execution state, we further introduce historical action conditioning and time-misalignment training, which provide short-horizon execution context and improve feedback control robustness under stale semantic conditions. Experiments on LIBERO with $\pi_{0.5}$ and UniVLA, together with real-robot deployment using UniVLA, show that the proposed framework achieves up to 35.6 Hz server-side action-module inference throughput and offers a low-intrusion path to high-frequency closed-loop control without running full VLA inference at control rate.
\end{abstract}



\begin{figure*}[t]
    \centering
    \includegraphics[width=0.98\textwidth]{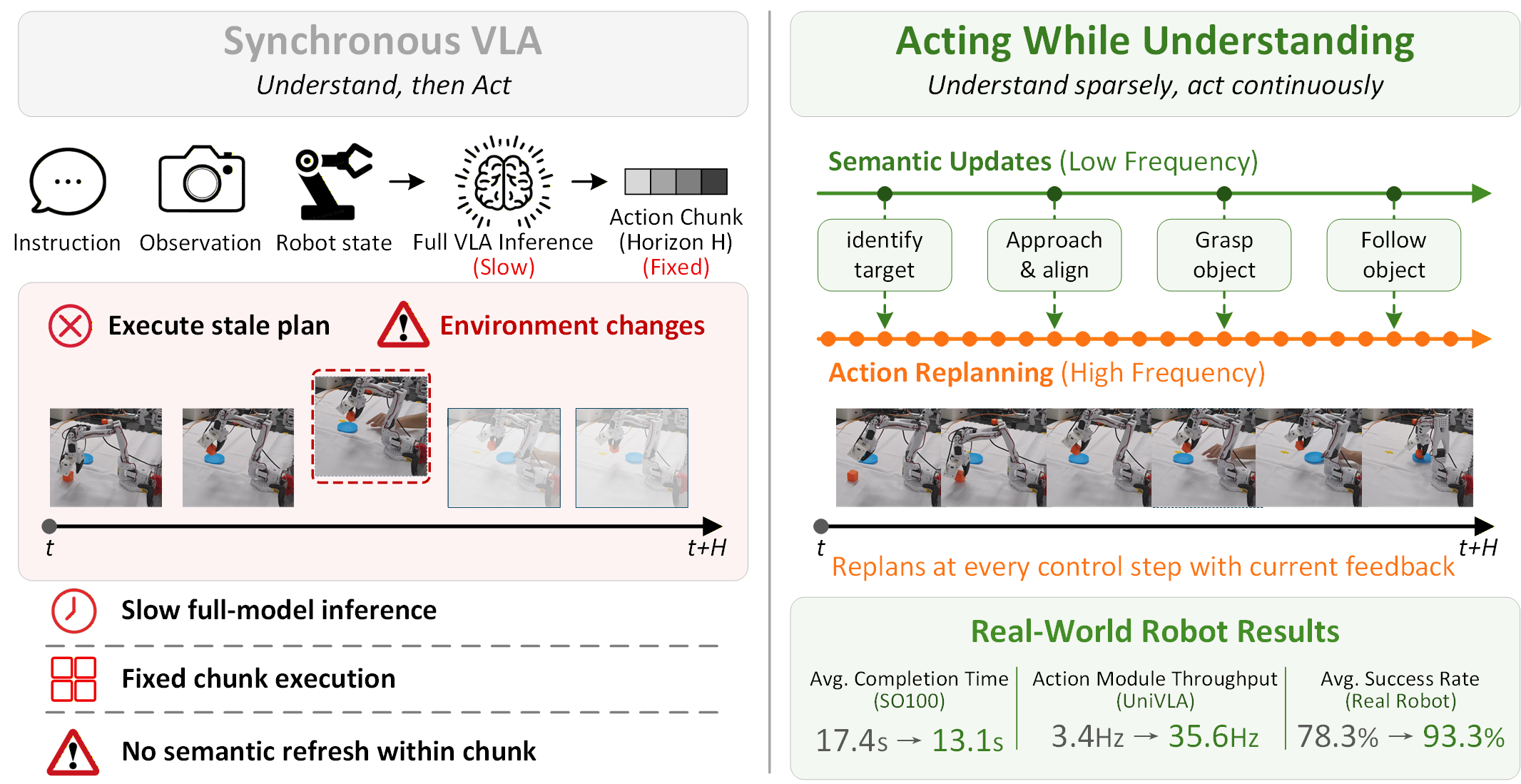}
    \caption{Comparison between conventional chunk-wise VLA execution and our asynchronous semantic-action decoupling. Conventional VLAs refresh semantics only at chunk boundaries, whereas our method refreshes semantics sparsely while continuously replanning actions from the latest semantic cache, current state, and action history.}
    \label{fig:teaser}
\end{figure*}


\section{Introduction}

Vision-Language-Action models (VLAs) are becoming an important class of policies for robotic manipulation. Compared with conventional task-specific imitation learning methods such as ACT~\cite{zhao2023ACT} and Diffusion Policy~\cite{chi2025DP}, VLA models jointly use visual observations, language instructions, and robot states for action generation, thereby exhibiting stronger potential in task understanding, cross-scene generalization, and long-horizon manipulation. Recent robot foundation models and VLAs, including RT-1, RT-2, PaLM-E, Open X-Embodiment / RT-X, Octo, RoboCat, vision-language foundation models as robot imitators, OpenVLA, SmolVLA, TinyVLA, Gemini Robotics, GR00T N1, $\pi_0$, $\pi_{0.5}$, and UniVLA, have substantially improved semantic reasoning, cross-task generalization, and cross-embodiment manipulation~\cite{brohan2022rt,zitkovich2023rt,driess2023palm,o2024open,team2024octo,bousmalis2023robocat,li2024vision,kim2024openvla,shukor2025smolvla,wen2025tinyvla,team2025gemini,bjorck2025gr00t,black2024pi0,intelligence2025pi05,bu2025univla}. These advances have also been supported by large-scale embodied datasets and benchmarks such as Open X-Embodiment, BridgeData V2, DROID, RLBench, CALVIN, RoboCasa, and LIBERO~\cite{o2024open,walke2023bridgedata,khazatsky2024droid,james2020rlbench,mees2022calvin,nasiriany2024robocasa,liu2023libero}. However, real-world robot control usually requires low-latency, high-frequency closed-loop action outputs, whereas full VLA inference often depends on computationally expensive vision-language backbones and action decoding. As a result, coordinating low-frequency semantic understanding with high-frequency action control has become a key challenge in deploying large VLAs on real robots.

Earlier language-conditioned and imitation-based manipulation systems such as SayCan, BC-Z, CLIPort, PerAct, RVT, Transporter Networks, Behavior Transformer, and Implicit Behavioral Cloning also established important ingredients for grounded manipulation and closed-loop visuomotor control~\cite{ahn2022can,jang2022bc,shridhar2022cliport,shridhar2023perceiver,goyal2023rvt,zeng2021transporter,shafiullah2022behavior,florence2022implicit}.

Existing VLA systems commonly adopt action chunking to mitigate the mismatch between full-model inference frequency and robot control frequency: a single inference step generates an action sequence over a future horizon, and a lower-level controller executes it at a fixed frequency. Although this strategy increases action density on the execution side and maintains continuous motion, it does not fundamentally remove the synchronous coupling between semantic understanding and action control. During chunk execution, the model's semantic judgment about the current task, goal, and scene usually remains unchanged; therefore, when the environment changes or task execution has already progressed, the robot may still rely on semantic and action intentions generated at an earlier time. Prior studies have attempted to alleviate this issue through model inference acceleration~\cite{xu2025vla,wang2025specvla}, action chunk continuation and correction~\cite{black2026real,sendai2025a2c2}, and asynchronous inference or slow-fast control~\cite{tang2025vlash,sun2026tidal}. Many of these methods, however, still rely on future-state compensation, diffusion-specific micro-loops, action-stream reordering, or additional execution modules. In contrast, we focus on a more direct question: under reusable but potentially stale semantic conditions, can we decouple action generation from full forward inference with minimal modification to existing VLAs while preserving high-frequency state-feedback control?

\begin{figure}[!t]
    \centering
    \includegraphics[width=0.98\linewidth]{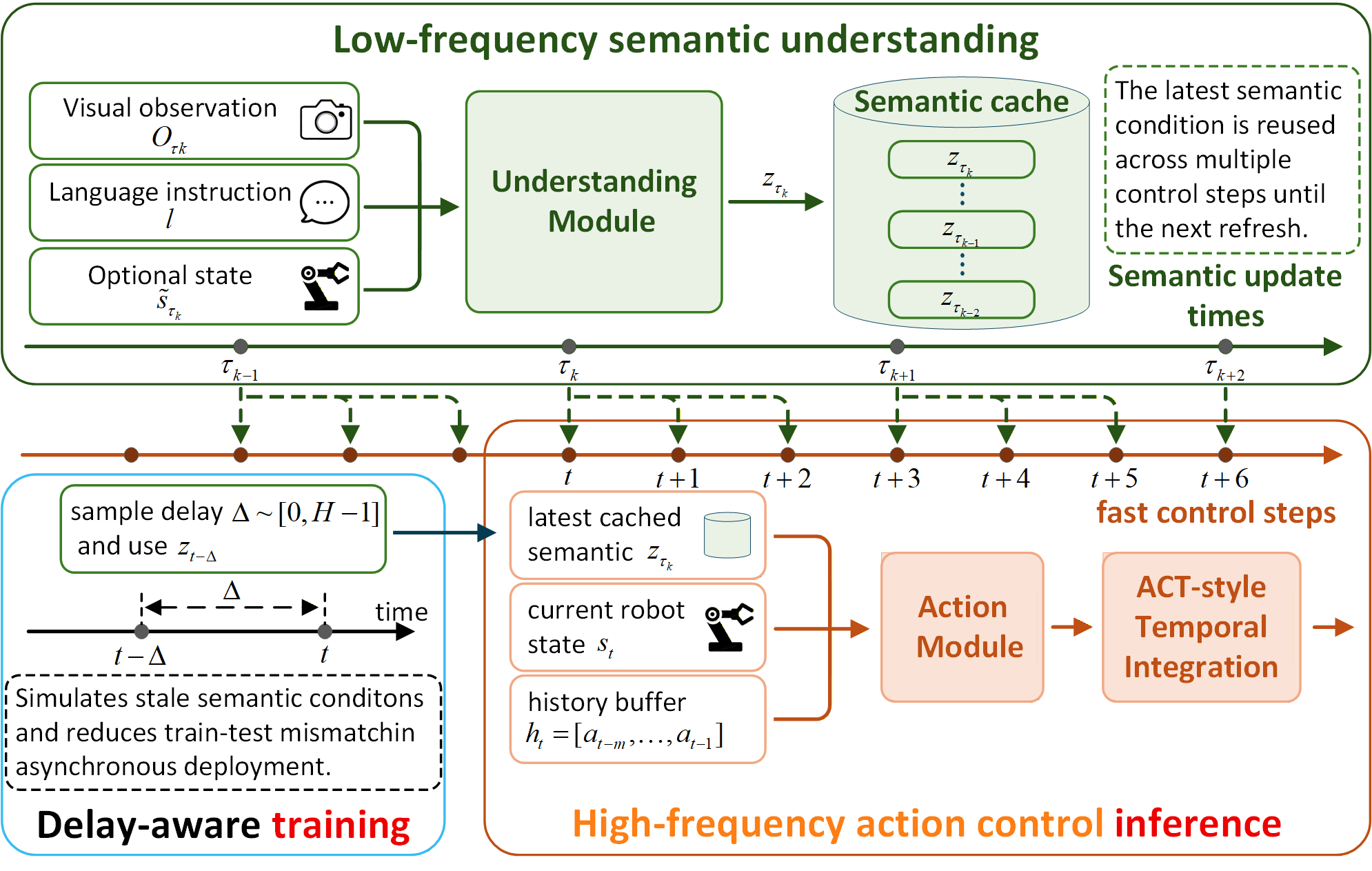}
    \caption{Overview of the proposed asynchronous semantic-action decoupling framework. A low-frequency understanding module asynchronously refreshes a semantic cache, while a high-frequency action module replans from the latest cached semantic condition, current robot state, and recent action history; delay-aware training exposes the action module to stale semantics, and ACT-style temporal integration merges overlapping replanned predictions.}
    \label{fig:async_framework}
    \vspace{-10pt}
\end{figure}
\vspace{-2pt}

To this end, we propose an asynchronous semantic-action decoupling framework that abstracts an existing VLA into two asynchronous execution roles: a low-frequency understanding module that processes visual observations and language instructions and updates cached semantic conditions, and a high-frequency action module that continuously generates actions conditioned on the latest available semantics, the current robot state, and recent execution history. Because cached semantics are often stale at deployment time, we further introduce historical action conditioning and time-misalignment training to mitigate the mismatch between outdated semantics and the current control state.

Our framework differs from prior asynchronous VLA systems in three aspects. First, we decouple along the \emph{internal semantic-action interface} of an existing VLA, rather than operating on the output action chunk stream after full-model inference. Second, the reused object is an intermediate \emph{semantic condition}---such as a last-layer hidden state or KV-cache---rather than a future action sequence to be continued, reordered, or corrected. Third, our main challenge is not merely how to reuse cached semantics, but how to maintain stable control when those semantics become stale; this is addressed by combining current-state feedback with historical action conditioning and explicit time-misalignment training.

We validate the proposed method in both simulation and real-robot settings. In LIBERO, we instantiate asynchronous variants of both $\pi_{0.5}$ and UniVLA to evaluate cross-backbone applicability. We further deploy UniVLA on real robots to verify practical closed-loop value. Experimental results show that the framework maintains strong semantic guidance while achieving up to 35.6 Hz server-side action-module inference throughput, enabling high-frequency closed-loop control without running full VLA inference at control rate.

Our contributions are threefold. First, we propose a low-intrusion asynchronous semantic-action decoupling framework along the internal semantic-action interface of pre-trained VLAs, letting a heavyweight understanding module run at low frequency while an action module runs at control rate. Second, we introduce historical action conditioning and time-misalignment training to improve robustness under stale semantic conditions. Third, we validate the framework on $\pi_{0.5}$, UniVLA, and real robots.


\section{Related Work}

\noindent{\bfseries Efficient VLA Inference and Real-Time Action Chunk Execution.}\quad
As VLA models scale up, reducing inference cost and improving execution continuity have become central deployment challenges. Existing methods improve efficiency through caching, speculative decoding, scheduling, pruning, routing, and sparsification~\cite{xu2025vla,wang2025specvla,li2025spvla,li2026cogvla,jiang2025asyncvla,jiang2026vla_perf}, while another line of work studies how to continue, correct, or reorder predicted action chunks during execution~\cite{black2026real,black2025training,liu2026legato,sendai2025a2c2,zhao2025vlarail,lu2026faster}. Beyond VLA-specific acceleration, practical deployment also benefits from strong vision-language backbones and efficient adaptation / serving techniques such as CLIP, Flamingo, DINOv2, PaliGemma, LoRA, QLoRA, FlashAttention, PagedAttention, and FAST~\cite{radford2021learning,alayrac2022flamingo,oquab2023dinov2,beyer2024paligemma,hu2022lora,dettmers2023qlora,dao2022flashattention,kwon2023efficient,pertsch2025fast}. These approaches alleviate the mismatch between full-model inference and control frequency, but they still largely center on full-model calls, output action-stream processing, or extra execution modules. In contrast, we target the temporal mismatch between semantic understanding and action generation inside existing VLAs, allowing the action module to run independently at high frequency from cached semantic conditions.

\noindent{\bfseries Asynchronous and Slow-Fast VLA Control.}\quad
The closest direction is asynchronous VLA execution and slow-fast robotic control, which separate high-level reasoning from fast motor updates~\cite{zhang2024hirt,chen2025fastinslow,lin2025onetwovla,xue2025reactive,tang2025vlash,sun2026tidal,xie2026dynamicvla}. Representative systems such as HiRT, Fast-in-Slow, OneTwoVLA, VLASH, TIDAL, and DynamicVLA explore this direction from hierarchical, dual-system, or asynchronous execution perspectives. These works recognize the multi-timescale nature of robotic control, but typically operate through hierarchical loops, future-state-aware inference, or action streaming. Our method instead decouples at the \emph{internal semantic-action interface}, reuses intermediate semantic conditions rather than future action sequences, and explicitly trains the action side to tolerate stale semantics through history conditioning and delay simulation.

\noindent{\bfseries Language-Conditioned Manipulation and Generalist Imitation Learning.}\quad
Before recent VLAs, grounded robot control was advanced by language-conditioned imitation and visuomotor policies such as SayCan, BC-Z, CLIPort, PerAct, RVT, and Transporter Networks~\cite{ahn2022can,jang2022bc,shridhar2022cliport,shridhar2023perceiver,goyal2023rvt,zeng2021transporter}. These methods connect semantic cues to closed-loop manipulation through affordance grounding, spatial reasoning, or structured visual representations. Behavior Transformer and Implicit Behavioral Cloning further show that strong sequence modeling and multimodal action learning from demonstrations can substantially improve policy expressiveness~\cite{shafiullah2022behavior,florence2022implicit}. These works establish important foundations for grounded manipulation and feedback control, but they typically do not separate semantic reasoning from action generation inside a single pre-trained VLA or explicitly study stale semantic reuse under asynchronous deployment.

\noindent{\bfseries Generalist Robot Data, Policies, and Benchmarks.}\quad
Recent progress in robot foundation models has also been enabled by broader datasets and evaluation suites. Open X-Embodiment / RT-X, BridgeData V2, and DROID expand data scale and embodiment diversity~\cite{o2024open,walke2023bridgedata,khazatsky2024droid}, while Octo, RoboCat, and SmolVLA explore increasingly accessible generalist robot policies~\cite{team2024octo,bousmalis2023robocat,shukor2025smolvla}. Benchmarks such as RLBench, CALVIN, RoboCasa, and LIBERO provide complementary testbeds for long-horizon, language-conditioned, and transfer-oriented manipulation~\cite{james2020rlbench,mees2022calvin,nasiriany2024robocasa,liu2023libero}. Our work is complementary to this trend: rather than proposing a new dataset or base policy, we study how to deploy existing large VLAs more effectively in high-frequency closed-loop settings.


\section{Method}
\label{sec:method}

\begin{figure}[t]
    \centering
    \includegraphics[width=0.98\linewidth]{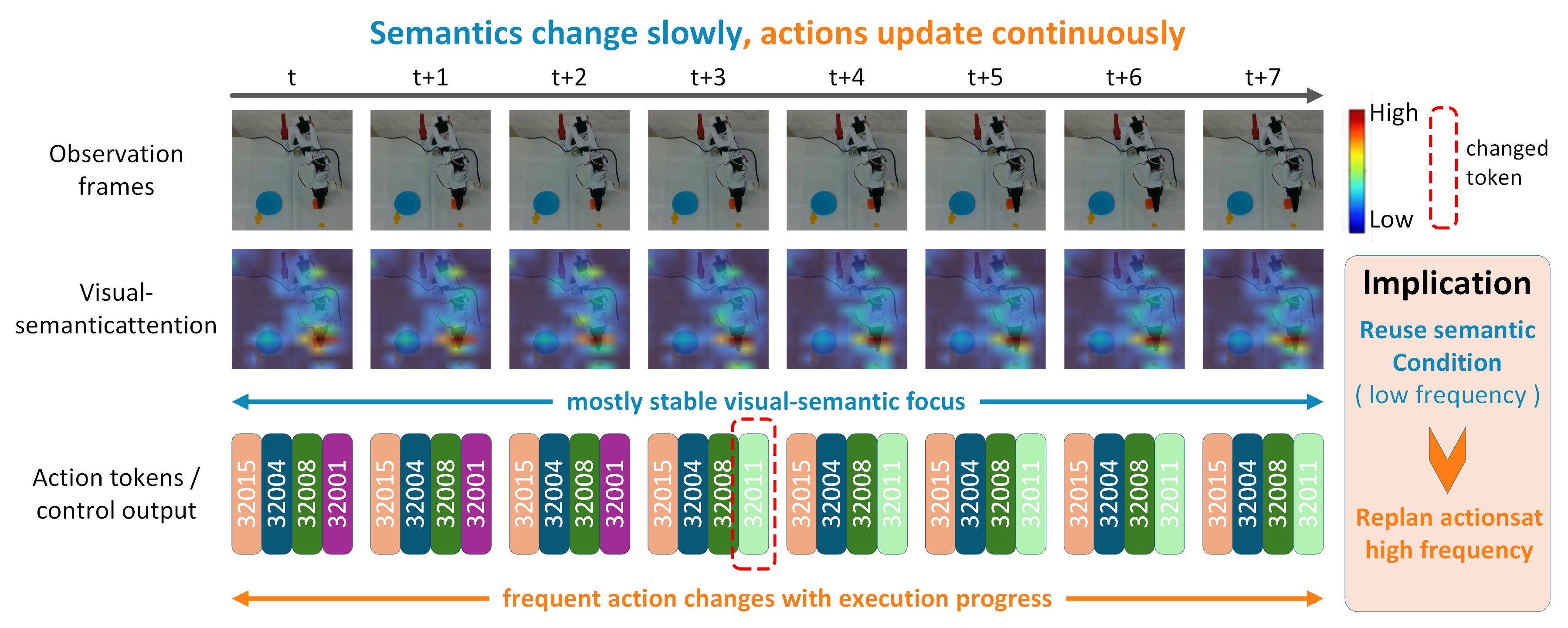}
    \caption{Temporal redundancy and action variation in consecutive VLA inference. Neighboring observations often share similar visual-semantic focus, while action predictions still vary with execution progress and local state, motivating high-frequency replanning under cached semantics.}
    \label{fig:temporal_redundancy}
\end{figure}

\subsection{Problem Formulation and Overview}

We start from the standard VLA formulation, where a policy predicts actions from current observations, language instructions, and optional robot state. Denoting by $U_\theta$ the semantic understanding computation and by $G_\phi$ the action generation computation, we abstract the policy as $z_t=U_\theta(o_t,l,\tilde{s}_t)$ and $A_t=G_\phi(z_t,s_t)$, where $z_t$ is a reusable intermediate condition before action generation and $\tilde{s}_t$ denotes optional state input to the understanding side. Here, the understanding and action modules refer to asynchronous functional roles rather than MoE-style experts or independently trained subnetworks. Unless otherwise specified, we inherit the original training protocol and action learning objective of the base VLA, without introducing an external planner or a complex multi-stage training pipeline.

At deployment time, $z_t$ primarily carries visual-language conditions, while real-time robot state feedback is always read by the action module at the current step. Concretely, we instantiate $z_t$ as the reusable condition immediately after visual-language grounding and before action generation: it is the last-layer VLM hidden state for UniVLA and the full cached KV-cache for $\pi_{0.5}$.

\subsection{Asynchronous Semantic-Action Decoupling}

The core observation behind asynchronous semantic-action decoupling is that high-level semantic understanding and low-level action control naturally evolve on different time scales: the former changes relatively slowly, while the latter requires frequent updates. As illustrated in Fig.~\ref{fig:temporal_redundancy}, neighboring frames in continuous control often exhibit only limited change in observations and attention focus, whereas substantial action adjustments are more likely to occur when the execution phase or target semantics changes. This suggests that intermediate semantic conditions can be reused across multiple control steps.

Based on this observation, we decouple a synchronous VLA into a low-frequency understanding module and a high-frequency action module. Let $\tau_k$ denote the time at which the $k$-th semantic update is completed and committed to the cache. The asynchronously updated semantic condition and the corresponding naive high-frequency action prediction can be written as
\begin{equation}
z_{\tau_k}=U_\theta(o_{\tau_k},l,\tilde{s}_{\tau_k}), \qquad \hat{A}_t=G_\phi(z_{\tau_k},s_t), \qquad \tau_k \le t.
\end{equation}
That is, at every control step the action module reads the latest committed semantic condition instead of waiting for a new full understanding result. When a new semantic condition arrives, subsequent action predictions automatically switch to the new condition without interrupting the current control loop. This formulation already removes the need to run the full VLA end-to-end at robot control frequency. However, relying only on the latest cached semantic condition and the current state still leaves progress ambiguity when the semantic condition becomes stale. We therefore treat this as a naive asynchronous reuse baseline, and then extend it with additional temporal context.

\subsection{Historical Action-Conditioned Action Module}

Although asynchronous decoupling allows the action module to run at high frequency under the latest available semantic condition, the semantic representation it reads may come from an earlier time step. Therefore, relying only on cached semantics and the current state is often insufficient to accurately reflect the current execution progress and local motion context. In continuous control, even when the high-level task semantics remain unchanged, the relative pose of the end-effector, contact state, and short-horizon motion trend continue to evolve over time.

To alleviate this temporal misalignment, we provide the action module with recent historical actions. Let $h_t=[a_{t-m},\ldots,a_{t-1}]$ denote a history window of length $m$. The full action generation process is then written as
\begin{equation}
\hat{A}_t = G_\phi(z_{\tau_k}, s_t, h_t), \qquad \tau_k \le t.
\end{equation}
In practice, we inject historical actions through a lightweight conditioning pathway on the action-generation side: UniVLA uses an additional projection layer, whereas $\pi_{0.5}$ introduces historical actions as an extra condition to the flow-matching action head. Cached semantics provide high-level task goals, while $h_t$ supplies recent execution trajectories and local temporal context that helps disambiguate execution progress under stale semantics. The understanding module and its output interface remain unchanged, so structural modifications are confined to the action-generation side.

\begin{wraptable}{r}{0.52\textwidth}
\vspace{-10pt}
\centering
\footnotesize
\setlength{\tabcolsep}{3.5pt}
\renewcommand{\arraystretch}{0.9}

\begin{tabular}{@{}lccccc@{}}
\toprule
Model & Spatial & Object & Goal & Long & Avg. \\
\midrule
$\pi_0$       & 96.8 & 98.8 & 95.8 & 85.2 & 94.2 \\
$\pi_{0.5}$  & 98.8 & 98.2 & 98.0 & 92.4 & 96.9 \\
UniVLA        & 96.5 & 96.8 & 95.6 & 92.0 & 95.2 \\
\midrule
$\pi_{0.5}$-15k       & 97.2 & 98.4 & 97.2 & 91.6 & 96.1 \\
UniVLA-Async          & 95.2 & 94.0 & 94.8 & 91.2 & 93.8 \\
$\pi_{0.5}$-Async-15k & \textbf{99.2} & 98.0 & \textbf{99.0} & \textbf{96.0} & \textbf{98.1} \\
\bottomrule
\end{tabular}

\vspace{-4pt}
\caption{
Main LIBERO results. The 15k-step $\pi_{0.5}$ variants use a reduced training budget relative to the publicly reported full-training baseline.
}
\label{tab:libero_main}
\vspace{-8pt}
\end{wraptable}

\subsection{Training and Inference}

If training always uses semantically synchronized conditions aligned with the current action, the action module may struggle at deployment time when semantic conditions become stale. We therefore explicitly simulate temporal misalignment during training. For action prediction at time $t$, instead of always conditioning on $z_t$, we use a stale semantic condition $z_{t-\Delta}$ from an earlier time, where $\Delta$ is sampled from $\{0,\ldots,H-1\}$. Here, $H$ is the action chunk length; in our setup, one chunk typically spans about $1\,\mathrm{s}$ of future control, so this sampling approximates a stale semantic delay of up to one action chunk. The action module is trained to learn
\begin{equation}
\hat{A}_t = G_\phi(z_{t-\Delta}, s_t, h_t),
\end{equation}
under the base action learning objective
\begin{equation}
\mathcal{L}=\mathbb{E}_{t,\Delta}\big[\ell_{\text{base}}(G_\phi(z_{t-\Delta}, s_t, h_t), A_t)\big],
\end{equation}
where $\ell_{\text{base}}$ denotes the original training objective of the backbone VLA, which may correspond to token prediction, flow matching, or action regression depending on the model. All experiments follow the official LIBERO fine-tuning configurations of the corresponding backbones, use the same datasets, and keep the default chunk length / action horizon settings. During training, $h_t$ is constructed from historical actions in the trajectory and perturbed with small Gaussian noise to approximate the mismatch between executed and expert actions; during inference, $h_t$ is built from actually executed actions.

At inference time, the understanding module asynchronously refreshes the semantic cache at a lower frequency, while the action module runs at every control step using the latest available semantic condition, current robot state, and recent action history. At each control step, the action module replans an action chunk, and ACT-style temporal integration combines temporally overlapping predictions into the executed action. In this way, the system no longer needs to run the full VLA at control frequency while still maintaining high-frequency state-feedback control under stale semantic conditions.


\section{Experiments}

We evaluate the proposed framework from three perspectives: simulation, real-robot deployment, and ablation analysis. We ask whether it preserves strong task success across backbones, improves real closed-loop behavior, and whether historical action conditioning and time-misalignment training are necessary under stale semantics.

\subsection{Experimental Setup}

Simulation experiments use LIBERO~\cite{liu2023libero} with asynchronous variants of both $\pi_{0.5}$ and UniVLA; task success rate is the primary metric. We report the official 96.9 $\pi_{0.5}$ result and use our $\pi_{0.5}$-15k variants as reduced-budget references. Since asynchronous $\pi_{0.5}$ is not deployed on our current robot setup, the real-world study focuses on synchronous versus asynchronous UniVLA.

Real-robot experiments use SO100 and Kinova Gen2, with three tasks per platform and 10 trials per task. The controlled real-world comparison is UniVLA versus UniVLA-Async under the same backbone, data, hardware, and action horizon; SmolVLA~\cite{shukor2025smolvla}, A2C2~\cite{sendai2025a2c2}, and synchronous $\pi_{0.5}$ serve as reference baselines. We report success rate, weighted SO100 completion time, and server-side inference throughput under the same A100 setting, while the robot control loop is fixed at 20 Hz. Since asynchronous $\pi_{0.5}$ is not deployed on our current robot setup, we report only its server-side throughput.

\subsection{Main Results on LIBERO}

We first report the main LIBERO results for asynchronous variants of UniVLA and $\pi_{0.5}$.

Table~\ref{tab:libero_main} shows that asynchronous semantic-action decoupling works across both UniVLA and $\pi_{0.5}$. For UniVLA, replacing synchronized semantic refresh with stale semantic reuse yields the expected moderate accuracy-real-time trade-off. $\pi_{0.5}$-Async remains highly competitive under only 15k training steps; because the official $\pi_{0.5}$ baseline uses a different training budget, we interpret this as evidence of non-degradation rather than a direct improvement. We conjecture that its robustness partly comes from reusing the full KV-cache rather than only the final hidden state before action generation.

\subsection{Real-World Experiments}

We further evaluate the deployability and cross-embodiment applicability of the framework on real robots. The core real-world comparison is UniVLA versus UniVLA-Async, while synchronous $\pi_{0.5}$ serves as a strong reference baseline under standard chunk-wise execution.

\begin{table}[t]
\centering
\footnotesize
\setlength{\tabcolsep}{4.0pt}
\renewcommand{\arraystretch}{1.02}

\begin{tabular*}{0.95\textwidth}{@{\extracolsep{\fill}}lccccccc@{}}
\toprule
\multirow{2}{*}{Method}
& \multicolumn{3}{c}{Kinova Gen2}
& \multicolumn{3}{c}{SO100}
& \multirow{2}{*}{Avg.} \\
\cmidrule(lr){2-4}
\cmidrule(lr){5-7}
& \makecell{pick\\banana}
& \makecell{stack\\cube}
& \makecell{pour\\water}
& \makecell{pick\\cube move}
& \makecell{stack\\cube}
& \makecell{open\\drawer}
& \\
\midrule
SmolVLA      & 70.0 & 70.0 & 30.0 & 70.0 & 70.0 & 80.0 & 65.0 \\
A2C2         & 40.0 & 50.0 &  0.0 & 40.0 & 60.0 & 50.0 & 40.0 \\
UniVLA       & 80.0 & 90.0 & 70.0 & 70.0 & 80.0 & 80.0 & 78.3 \\
UniVLA-Async & 90.0 & 100.0 & 90.0 & 100.0 & 90.0 & 90.0 & 93.3 \\
$\pi_{0.5}$  & 100.0 & 100.0 & 90.0 & 90.0 & 100.0 & 90.0 & 95.0 \\
\bottomrule
\end{tabular*}

\vspace{2pt}
\caption{
Main real-robot results. We report task-level success rates (\%) over 10 trials and the average success rate across six tasks.
}
\label{tab:real_success}
\end{table}

Table~\ref{tab:real_success} reports the main real-world success rates across the six tasks.

Table~\ref{tab:real_runtime} further characterizes runtime behavior in real closed-loop execution. On SO100, UniVLA-Async reduces the weighted average completion time from 17.4s to 13.1s, indicating that the benefit of asynchronous deployment appears not only in success rate but also in more continuous execution. Synchronous UniVLA reaches 3.4 Hz server-side throughput, UniVLA-Async reaches 35.6 Hz, and asynchronous $\pi_{0.5}$ reaches 19.7 Hz; by contrast, A2C2 is faster but substantially less successful, indicating that throughput alone is insufficient. On Kinova, timing is less discriminative because physical joint speed limits dominate the overall episode duration, so we focus primarily on success rate.

\begin{wrapfigure}{r}{0.48\linewidth}
\vspace{-10pt}
\centering
\includegraphics[width=\linewidth]{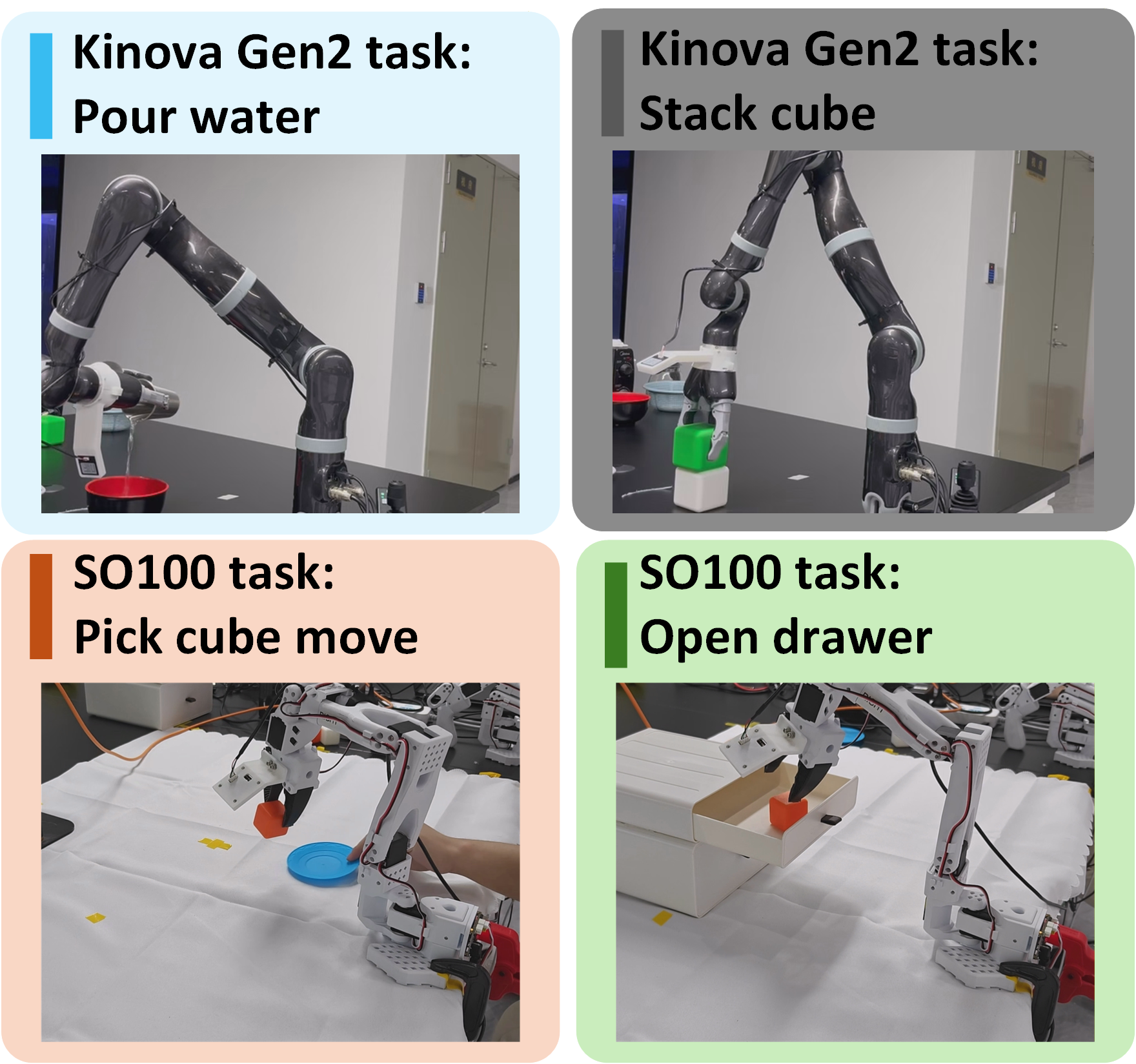}
\caption{
Representative real-world task snapshots on Kinova Gen2 and SO100.
}
\label{fig:real_exp}
\vspace{-10pt}
\end{wrapfigure}

\subsection{Ablation on Stale Semantic Robustness}

To verify that our method is not merely simple cache reuse, we perform an ablation study centered on stale semantic conditions. Because semantic delay is most pronounced in long-horizon tasks, we focus on UniVLA on LIBERO-Long and compare the naive asynchronous reuse baseline, history-only conditioning, delay-only training, and the full method.

\begin{table}[t]
\vspace{-6pt}
\centering
\footnotesize
\setlength{\tabcolsep}{4pt}
\renewcommand{\arraystretch}{0.95}

\begin{minipage}[t]{0.52\linewidth}
\vspace{0pt}
\centering
\begin{tabular}{@{}lcc@{}}
\toprule
Method & \makecell{SO100 Avg.\\Time (s)} & \makecell{Throughput\\(Hz)} \\
\midrule
SmolVLA      & 15.7 & 3.9 \\
A2C2         & 20.3 & 61.6 \\
UniVLA       & 17.4 & 3.4 \\
UniVLA-Async & \textbf{13.1} & 35.6 \\
$\pi_{0.5}$  & 14.6 & 8.7 \\
$\pi_{0.5}$-Async & -- & 19.7 \\
\bottomrule
\end{tabular}
\captionof{table}{Runtime summary on real robots. SO100 completion time and server-side throughput.}
\label{tab:real_runtime}
\end{minipage}
\hfill
\begin{minipage}[t]{0.40\linewidth}
\vspace{0pt}
\centering
\begin{tabular}{@{}lc@{}}
\toprule
Method & LIBERO-Long \\
\midrule
Sync UniVLA          & 92.0 \\
Naive cache reuse    & 85.2 \\
Cache + history      & 86.4 \\
Cache + delay train. & 89.0 \\
\textbf{Ours}        & \textbf{91.2} \\
\bottomrule
\end{tabular}
\captionof{table}{Ablation on LIBERO-Long under stale semantic conditions.}
\label{tab:ablation_long}
\end{minipage}
\vspace{-8pt}
\end{table}

Table~\ref{tab:ablation_long} shows that directly replacing synchronized semantics with stale cached semantics leads to clear degradation: naive cache reuse drops from 92.0 to 85.2, indicating that reusing cached semantics alone is insufficient for stable control. Adding only historical action conditioning recovers performance to 86.4, while adding only delay training reaches 89.0, suggesting that both mechanisms help mitigate errors induced by stale semantics, with explicit modeling of train-test temporal misalignment being especially important. Combining both historical action conditioning and time-misalignment training raises performance to 91.2, close to the synchronous UniVLA score of 92.0. This shows that the key of the proposed method lies not in caching itself, but in enabling the action module to control stably under stale semantic conditions by combining current-state feedback with recent execution context.


\section{Limitations and Discussion}

Although the proposed method performs well in both simulation and real-robot experiments, it still has several limitations. First, the framework depends on the existence of a reusable intermediate condition before action generation, so its applicability remains tied to the internal interface form of the base VLA. Second, although historical action conditioning and time-misalignment training alleviate performance degradation under stale semantic conditions, the action module may still be limited when semantic cache age becomes too large or when the environment changes significantly. In particular, the current fixed semantic refresh schedule is not always optimal: when scene changes, execution phase transitions, or uncertainty spikes occur rapidly, stale semantics may still mislead the controller before the next refresh. Finally, our experiments are mainly concentrated on single-arm manipulation tasks; although they cover long-horizon, multi-stage, and perturbation-sensitive scenarios, we have not yet validated the method on fast dynamic manipulation, mobile manipulation, or more complex contact-rich interactions.

Future work includes adaptive semantic refresh and broader validation across VLA architectures and real-world tasks. Overall, our results suggest that low-intrusion asynchronous decoupling is a practical path toward deploying large VLAs for real-time robot control.


\clearpage

\bibliography{references}

\clearpage
\appendix
\section{Implementation Details}
\label{app:implementation}

This appendix summarizes implementation details omitted from the main paper, including the concrete instantiation of the semantic-action interface, the way historical actions are injected, the training protocol inherited from each backbone, and the asynchronous deployment procedure used at inference time.

\subsection{Instantiating the Semantic-Action Interface}

For UniVLA, we follow the original design and instantiate the reusable semantic condition $z_t$ using the last-layer hidden states on the vision and additional output tokens, excluding the prompt tokens. These hidden states are then passed through the original projection interface before entering the action head. In the real-robot UniVLA setup, the current robot state is additionally re-introduced on the action side, since the original UniVLA action head does not explicitly read state input. Concretely, this state input consists of the current joint state together with the gripper state. In the LIBERO simulation experiments, by contrast, we follow the original UniVLA simulation setup without this extra action-side state input.

For $\pi_{0.5}$, the reusable semantic condition is instantiated as the vision-language-conditioned prefix KV-cache. During asynchronous deployment, this cache is reused by the action-generation side, while the architectural modification is confined to the action head. This design follows the central principle of our framework: decoupling is introduced at the semantic-action interface with minimal intrusion to the original backbone.

Across both backbones, the language instruction is fixed during inference. Visual observations are refreshed only by the low-frequency understanding module. The action module itself does not perform a new full visual-language forward pass, but instead reuses the latest available semantic condition provided by the understanding side.

\subsection{Historical Action Injection}

We provide the action module with a short history of previously executed actions to compensate for the temporal mismatch between stale semantic conditions and the current execution phase. The history length is chosen to match approximately one second of control, following the chunking configuration of each backbone. In LIBERO, this corresponds to a history length of 8 for UniVLA and 10 for $\pi_{0.5}$. In real-robot deployment, the history window is set to 20 control steps, matching approximately one second under the 20 Hz control loop.

The history is represented in the same action space as the policy output. That is, end-effector policies use end-effector action histories, joint-space policies use joint-space action histories, and gripper actions are included when present. We do not apply extra normalization beyond the backbone's original action representation.

For UniVLA, the historical actions are mapped through a lightweight projection layer and then injected into the action-generation side as an additional conditioning input. For $\pi_{0.5}$, the historical actions are introduced as an extra condition embedding for the action head. In both cases, the understanding module remains unchanged, and the modification is restricted to the action side.

During training, the action history is constructed from ground-truth trajectory actions with small Gaussian perturbations to better approximate the mismatch between executed and demonstrated actions. During inference, the history is maintained online from the actually executed actions.

\subsection{Training Details}

Unless otherwise specified, all training protocols follow the official backbone configurations. For UniVLA, we follow the original LIBERO training recipe: the VLM is adapted with LoRA and then frozen, while the action expert is trained from scratch. In LIBERO simulation, UniVLA-based models are trained on 8 A100 GPUs with a global batch size of 128 for 30k steps on each of the four LIBERO suites.

For $\pi_{0.5}$, our asynchronous LIBERO experiments are conducted as reduced-budget runs. We fully fine-tune the model on 2 A100 GPUs with a global batch size of 64 for 15k steps. For reference, the official full-training result is obtained under a larger setup with 8 A100 GPUs, a global batch size of 256, and 30k training steps.

Time-misalignment training follows the formulation in the main text. Specifically, the semantic delay $\Delta$ is uniformly sampled over one action chunk, i.e., $\Delta \sim \mathrm{Uniform}\{0,\ldots,H-1\}$ where $H$ is the chunk size of the corresponding backbone. This setting is shared across both backbones. All methods use the same simulation splits and evaluation protocol, and final results are reported as averages over repeated evaluation runs.

\subsection{Inference and Temporal Integration}

During deployment, the understanding module and the action module run as asynchronous processes. The understanding side typically refreshes the semantic cache at about 3--5 Hz, and a new semantic update is started immediately after the previous one finishes. In contrast, the action module runs at every control step and replans a full action chunk using the latest available semantic condition, the current robot state, and the recent action history.

At each step, the replanned chunk is not executed directly as-is. Instead, we follow ACT-style temporal integration to merge temporally overlapping predictions and output the final action sent to the robot. The integration window follows the dataset FPS / chunking configuration, and we use the same temporal weighting rule in both simulation and real-robot deployment.

During asynchronous execution, the action side always reads the latest committed semantic cache rather than any in-progress semantic update. Newly produced semantic outputs become visible to the action module only after the corresponding semantic refresh is complete and committed. For temporal integration, we follow the original ACT-style exponential weighting rule with decay coefficient 0.1 in both simulation and real-robot deployment.

Our real-robot system uses a server-client architecture in which an A100 server performs model inference and the robot side maintains the control loop. Communication is handled over an Ethernet network using an HTTP(S)-based protocol. The model-throughput numbers reported in the main paper reflect pure server-side inference throughput and do not include communication overhead, whereas the real-world control loop naturally includes the end-to-end communication latency, which is typically around 10--15 ms.

\section{Additional Experimental Settings}
\label{app:settings}

This appendix provides additional settings for LIBERO and real-world experiments, including the inherited backbone configurations, reduced-budget training conditions, deployment protocol, and the fairness criteria used for comparison.

\subsection{LIBERO Setup and Training Protocol}

Unless otherwise specified, all LIBERO experiments use the four standard suites: LIBERO-Spatial, LIBERO-Object, LIBERO-Goal, and LIBERO-Long. We follow the original task splits and do not exclude or specially reprocess any tasks. Observation formatting, image inputs, and language instruction templates are inherited from the corresponding backbone implementations.

The asynchronous variants also keep the default action chunk / action horizon settings of their original backbones. In LIBERO, UniVLA uses a chunk size of 8, while $\pi_{0.5}$ uses a chunk size of 10. For UniVLA, the LIBERO simulation results follow the original simulation setup without extra action-side state input, whereas the real-robot experiments follow the original UniVLA real-robot convention that reintroduces joint state and gripper state on the action side.

Evaluation follows the standard suite-level success-rate protocol. Each suite contains 10 tasks, and each task is evaluated over 50 rollout episodes from the same checkpoint, resulting in 500 rollouts per suite. The reported success rate is computed at the suite level under this repeated-rollout protocol. Random seed handling is aligned with the OpenVLA / UniVLA evaluation environment so that all compared methods follow the same seed protocol.

In Table~\ref{tab:libero_main}, the 96.9 result of $\pi_{0.5}$ is the publicly reported number from the official repository. By contrast, $\pi_{0.5}$-15k and $\pi_{0.5}$-Async-15k are trained under a substantially reduced budget on two GPUs, compared with the official 8-GPU configuration. The 15k-step synchronous and asynchronous $\pi_{0.5}$ results therefore serve as matched reduced-budget references under the same evaluation protocol.

Accordingly, we do not interpret the reduced-budget asynchronous $\pi_{0.5}$ results as a direct improvement over the official full-training baseline. Instead, this comparison is intended to show that asynchronous semantic reuse remains competitive and does not introduce obvious degradation even under a smaller training budget.

\FloatBarrier

\subsection{Real-World Deployment Protocol}

Each real-world task uses a fixed language instruction across all compared methods. Trials are initialized with small random perturbations in object configuration, while the human perturbation protocol is applied only in the SO100 \texttt{pick\_cube\_move} task. After each trial, the scene is manually reset before the next rollout.

We evaluate six real-world tasks in total, three on SO100 and three on Kinova Gen2, with 10 trials per task. Task success is determined by manual judgment based on whether the target manipulation objective is completed. Concretely, \texttt{pick\_banana} requires moving the banana onto the target plate, \texttt{stack\_cube} requires a stable cube stack, \texttt{pour\_water} requires pouring the water into the bowl and returning the cup to its original place, \texttt{pick\_cube\_move} requires placing the cube onto the blue plate, and \texttt{open\_drawer} requires opening the drawer, placing the cube inside, and closing it again.

\begin{table*}[htbp]
\centering
\setlength{\tabcolsep}{4pt}
\renewcommand{\arraystretch}{1.05}
\footnotesize
\begin{tabular}{@{}llp{0.42\textwidth}p{0.23\textwidth}@{}}
\hline
Platform & Task & Description & Characteristic \\
\hline
Kinova & \texttt{pick\_banana} & Move the banana from the table to the silver plate. & Pick-and-place \\
Kinova & \texttt{stack\_cube} & Stack the green cube on top of the white cube. & Precision-sensitive stacking \\
Kinova & \texttt{pour\_water} & Pour the water from the metal cup into the bowl. & Long-horizon task with an extended waiting stage \\
SO100 & \texttt{pick\_cube\_move} & Move the orange cube to the blue plate under human perturbation. & Perturbation-sensitive pick-and-place \\
SO100 & \texttt{stack\_cube} & Stack the orange cube on top of the black cube. & Precision-sensitive stacking \\
SO100 & \texttt{open\_drawer} & Open the drawer, place the cube inside, then close it. & Multi-stage long-horizon manipulation \\
\hline
\end{tabular}
\caption{Task descriptions and key characteristics of the real-robot experiments.}
\label{tab:app_real_tasks}
\end{table*}

For SO100, completion time is measured from the onset of action execution until successful task completion, and failed trajectories are excluded from this timing statistic. The weighted average completion time reported in the main paper is computed by weighting each task according to its number of successful trajectories rather than by taking a simple arithmetic mean. Both robot platforms run with a fixed 20 Hz control loop.

Deployment follows a server-client setup, with an A100 server used for inference and a robot-side control loop fixed at 20 Hz. At every control step, the robot sends the latest state to the server-side action module, which replans an action chunk under the latest committed semantic condition and returns the temporally integrated action for execution. The average inference frequency reported in Table~\ref{tab:real_runtime} corresponds to the throughput of the server-side action module, rather than the actual command dispatch rate on the robot side. By contrast, real-world execution naturally includes communication latency, which is typically around 10--15 ms in our setup.

The Kinova Gen2 platform is additionally constrained by the official driver-side joint speed limits, specifically J1--J3 at 36 deg/s (0.628 rad/s) and J4--J6 at 48 deg/s (0.838 rad/s). These limits are part of the standard platform configuration rather than an extra restriction introduced by our system. As a result, physical arm motion dominates overall episode duration on Kinova, which is why completion time is less informative there than success rate.

\FloatBarrier

\subsection{Baseline Fairness and Evaluation Scope}

Across both simulation and real-robot experiments, all methods use the same task data and keep the default chunk size / action horizon under their official configurations. SmolVLA, A2C2, UniVLA, and synchronous $\pi_{0.5}$ all follow standard chunk-wise execution, whereas UniVLA-Async performs online replanning at each control step and applies temporal integration to generate the executed action. The most controlled real-world comparison is between UniVLA and UniVLA-Async, since these two methods share the same backbone, data, hardware, and action horizon. For fairness, the synchronous UniVLA baseline in this real-world comparison uses the same state-conditioned action-side interface as the asynchronous UniVLA setup, following the original UniVLA real-robot deployment convention; the main difference is whether semantic conditions are refreshed synchronously or reused asynchronously.

For $\pi_{0.5}$, the asynchronous variant is not deployed on the real robots. The reported 19.7 Hz figure reflects only server-side throughput rather than end-to-end real-robot closed-loop execution. This distinction is important in practice: after accounting for communication overhead, the throughput gain from asynchronous decoupling is more limited for the 2B $\pi_{0.5}$ model than for the 7B UniVLA model, and is not sufficient to stably track the full 20 Hz robot-side control loop in our current setup. By contrast, the asynchronous gain is substantially larger for UniVLA, where decoupling the heavier backbone produces a much larger practical benefit.

All throughput numbers reported in the main paper are measured under the same A100 server-side setting. The average inference frequency of A2C2 is measured according to its original corrective-head inference procedure; therefore, higher throughput does not necessarily imply better closed-loop performance, which is also reflected in its lower task success in the real-robot study.

Finally, the reduced-budget $\pi_{0.5}$ results should be interpreted carefully. Our conclusion is not that asynchronous $\pi_{0.5}$ directly outperforms the official full-budget baseline under a matched setup. Rather, the result suggests that asynchronous decoupling remains effective even under a much smaller training budget, likely benefiting from the strong pretrained foundation of $\pi_{0.5}$ and the richer intermediate information retained in its KV-cache representation. More broadly, our real-world study focuses on single-arm manipulation and does not yet cover mobile manipulation or highly dynamic tasks.

\FloatBarrier

\section{Additional Quantitative Results}
\label{app:quantitative}

This appendix reports additional quantitative results not expanded in the main paper, including task-level real-world descriptions, task-wise timing statistics, and clarifications on the runtime and ablation analyses.

\subsection{Detailed Real-World Results}

Table~\ref{tab:app_real_tasks} summarizes the task descriptions and key characteristics of the two real-robot platforms. In addition to the platform-level average success rates in the main text, Table~\ref{tab:app_so100_time} reports task-wise average completion times on SO100 over successful trajectories.

\begin{table}[htbp]
\centering
\setlength{\tabcolsep}{4pt}
\renewcommand{\arraystretch}{0.95}
\footnotesize
\begin{tabular}{lccccc}
\hline
Task & SmolVLA & A2C2 & UniVLA & UniVLA-Async & $\pi_{0.5}$ \\
\hline
\texttt{pick\_cube\_move} & 14.5 & 20.4 & 17.6 & 11.2 & 11.8 \\
\texttt{stack\_cube} & 10.5 & 14.6 & 11.1 & 9.2 & 12.8 \\
\texttt{open\_drawer} & 21.2 & 26.8 & 23.2 & 19.1 & 19.3 \\
\hline
\end{tabular}
\caption{Task-wise average completion time on SO100 over successful trajectories (seconds).}
\label{tab:app_so100_time}
\end{table}

\subsection{Additional Runtime Statistics}

Table~\ref{tab:real_runtime} already summarizes the weighted average completion time on SO100 and the average server-side inference throughput. Here the weighted average is computed according to the number of successful trajectories in each task, rather than as a simple arithmetic mean across tasks:
\begin{equation}
T_{\text{weighted}}=\frac{\sum_i n_i T_i}{\sum_i n_i},
\end{equation}
where $T_i$ is the average completion time of task $i$ over successful trajectories and $n_i$ is the number of successful trajectories for that task.

The SO100 task-wise timing statistics in Table~\ref{tab:app_so100_time} further show that the improvement of UniVLA-Async is consistent across perturbation-sensitive, precision-sensitive, and multi-stage tasks, rather than being driven by only a single category.

\begin{table*}[htbp]
\centering
\setlength{\tabcolsep}{5pt}
\renewcommand{\arraystretch}{1.02}
\footnotesize
\begin{tabular}{lccccc}
\hline
Metric & Mean & Std & P50 & P95 & Unit \\
\hline
Semantic module latency & 319.1 & 7.5 & 319.3 & 325.0 & ms \\
Action-side inference latency & 4.63 & 0.24 & 4.62 & 4.98 & ms \\
Policy step latency & 11.22 & 26.50 & 8.34 & 9.99 & ms \\
End-to-end control latency & 11.62 & 26.50 & 8.74 & 10.42 & ms \\
VLA enqueue delay & 45.0 & 53.4 & 33.5 & 69.9 & ms \\
Semantic age at control step & 163.9 & 93.0 & 163.1 & 308.3 & ms \\
Latent lag & 8.27 & 2.03 & 8.00 & 11.00 & frames \\
\hline
\end{tabular}
\caption{Additional runtime statistics of the asynchronous UniVLA deployment collected over a 100-step run. Semantic refresh is observed at 3.07 Hz in this run. The reported action-side inference latency includes action-decoder invocation overhead and CUDA synchronization, but excludes temporal integration, communication, and semantic-module execution.}
\label{tab:app_runtime_breakdown}
\end{table*}

Table~\ref{tab:app_runtime_breakdown} provides a more detailed runtime view of the asynchronous UniVLA deployment. The semantic module remains the dominant cost, with a mean latency of about 319 ms, whereas the action-side inference latency is about 4.6 ms on average. This separation is exactly what enables the asynchronous design: the heavy semantic refresh proceeds at low frequency, while the lightweight action side continues to replan at control rate using the most recent committed semantic condition. The action-side latency in Table~\ref{tab:app_runtime_breakdown} should not be interpreted as the inverse of the effective throughput reported in Table~\ref{tab:real_runtime}. The former measures the latency of one action-side inference invocation, including decoder-side overhead and CUDA synchronization, whereas the latter reflects the effective server-side action-generation frequency of the deployed asynchronous loop, which is additionally affected by queueing, scheduling, semantic-cache coordination, and other per-step system overheads outside the action-side timing window.

The relatively large standard deviation of policy-step latency and end-to-end control latency is caused by rare scheduling outliers beyond the 95th percentile, while the bulk of control steps remains concentrated near the reported median and 95th-percentile values.

We further note that the reported $\pi_{0.5}$-Async throughput reflects only server-side model throughput and does not correspond to a real-robot deployment result. This distinction is important because server-side throughput alone does not determine whether a method can stably follow the full 20 Hz robot-side control loop once communication overhead is included. In particular, although asynchronous decoupling improves the server-side throughput of the 2B $\pi_{0.5}$ model from 8.7 Hz to 19.7 Hz, the remaining communication and control overhead still makes stable 20 Hz closed-loop tracking less reliable in our current setup.

The absence of Kinova timing statistics is not due to missing data, but because the platform's strict joint speed limits amplify physical execution time to the point that inference-induced timing differences become much less discriminative in the total episode duration. For this reason, Kinova is used primarily to evaluate robustness and cross-platform deployability through success rate rather than completion time.

\subsection{Additional Ablation Clarifications}

The main paper focuses on the ablation results on LIBERO-Long in Table~\ref{tab:ablation_long}, because stale semantic conditions are most pronounced in long-horizon tasks. Among the four LIBERO suites, LIBERO-Long is relatively more difficult and yields consistently lower scores across methods, making it the most direct testbed for stale-semantic robustness.

The ablation is designed to isolate the contribution of the two key robustness components under stale semantics. ``Naive cache reuse'' directly replaces synchronized semantic refresh with stale cached semantics while keeping the rest of the policy unchanged. ``Cache + history'' additionally provides recent action history to the action module. ``Cache + delay train.'' instead keeps the asynchronous formulation but exposes the model to semantic delay during training. ``Ours'' combines both historical-action conditioning and time-misalignment training.

The resulting pattern is consistent with the main claim of the paper: naive semantic reuse alone is insufficient, while combining short-horizon execution context with explicit delay simulation recovers most of the lost performance. In particular, the stronger gain from delay training than from history-only conditioning suggests that train-test temporal mismatch is the central difficulty introduced by asynchronous semantic reuse, while historical actions act as a complementary local progress signal.

In the current version, we do not further expand ablations over different history lengths, semantic refresh intervals, or integration settings, and we do not include a separate $\pi_{0.5}$ ablation. This keeps the ablation section focused on the clearest question raised by the paper: whether the two proposed robustness mechanisms are both necessary when semantics become stale.

\section{Additional Qualitative Results}
\label{app:qualitative}

This appendix provides additional qualitative material for the real-robot experiments. Since static figures are better suited to illustrating task stages than to evaluating motion smoothness, we focus here on representative task progressions across platforms and reserve continuous motion comparisons for the supplementary videos.

\subsection{Real-World Rollout Visualizations}

\begin{figure*}[htbp]
\centering
\includegraphics[width=0.995\textwidth]{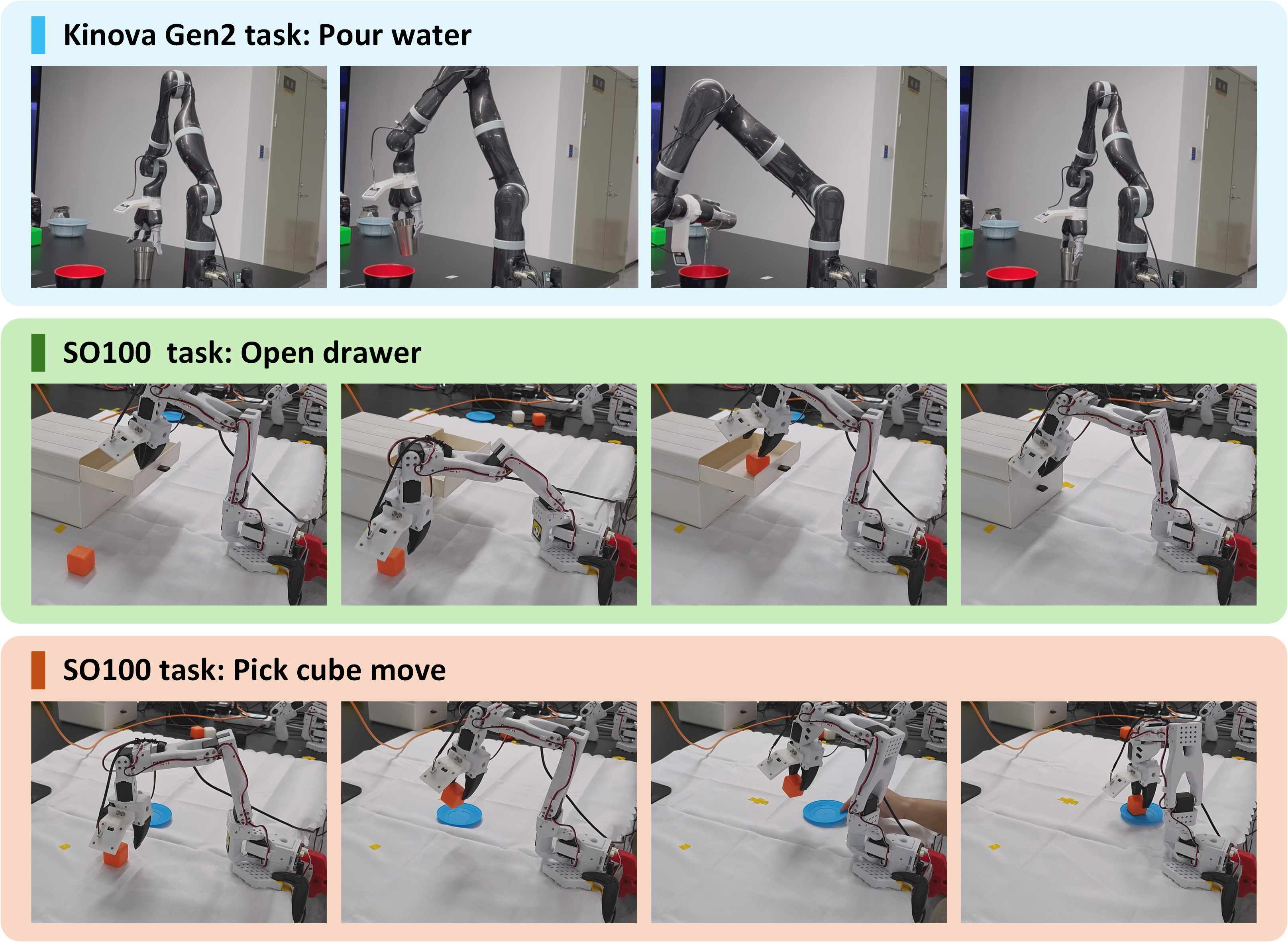}
\caption{Representative real-world task progressions on Kinova Gen2 and SO100. From top to bottom, the rows correspond to \texttt{pour\_water}, \texttt{open\_drawer}, and \texttt{pick\_cube\_move}. Within each row, frames are arranged from left to right in chronological order to illustrate key execution stages of the task. These static visualizations are intended to show representative task progression and deployment context; continuous motion smoothness and recovery behavior are better reflected in the supplementary videos.}
\label{fig:app_real_rollouts}
\end{figure*}

Figure~\ref{fig:app_real_rollouts} shows representative task progressions for three real-robot tasks: \texttt{pour\_water} on Kinova Gen2, and \texttt{open\_drawer} and \texttt{pick\_cube\_move} on SO100. These tasks are chosen because they cover three qualitatively different execution patterns in our study: long-horizon pouring, multi-stage object placement with drawer interaction, and perturbation-sensitive pick-and-place.

Within each row, the frames are ordered chronologically from left to right and illustrate the main stages of task execution. The goal of this figure is to make the task structure and deployment scenarios visually concrete, rather than to claim that static images alone can faithfully characterize motion continuity. In particular, \texttt{pick\_cube\_move} highlights the human-perturbation setting used on SO100, while \texttt{open\_drawer} and \texttt{pour\_water} emphasize multi-stage execution under longer manipulation horizons.

\FloatBarrier

\subsection{Additional Qualitative Observations}

Across both platforms, all six real-world tasks were recorded for qualitative inspection. The examples in Fig.~\ref{fig:app_real_rollouts} are selected as representative cases because they most clearly expose long-horizon execution, stage transitions, and perturbation-sensitive behavior. They therefore complement the quantitative results in the main paper by showing the concrete manipulation scenarios behind the reported success rates and completion times.

For the purposes of the appendix, we emphasize representative task progression rather than dense motion-level comparison. Static figures are sufficient for showing the sequence of sub-tasks and the overall deployment context, whereas subtle differences in action continuity, hesitation, or recovery are better conveyed through full rollout videos. Additional real-robot videos across all tasks and models can therefore be included as supplementary material when needed.

\end{document}